\documentclass[10pt,twocolumn,letterpaper]{article}

\def\compileArxiv{true}    

\usepackage{cvpr}              

\usepackage{algorithm}
\usepackage{algpseudocode}
\usepackage{booktabs}
\usepackage{multirow}
\usepackage{amssymb}

\usepackage{fancyhdr}
\usepackage{comment}

\newcommand{\cmark}{\checkmark}
\newcommand{\xmark}{\ensuremath{\times}}

\definecolor{cvprblue}{rgb}{0.21,0.49,0.74}
\usepackage[pagebackref,breaklinks,colorlinks,allcolors=cvprblue]{hyperref}

\usepackage[normalem]{ulem}
\def\paperID{3} 
\def\confName{CVPR}
\def\confYear{2026}

\title{Guideline2Graph: Profile-Aware Multimodal Parsing for Executable Clinical Decision Graphs}

\author{
Onur Selim Kilic$^{1}$ \quad
Yeti Z. Gurbuz$^{2}$ \quad
Cem O. Yaldiz$^{1}$ \quad
Afra Nawar$^{1}$ \quad
Etrit Haxholli$^{2}$\\
Ogul Can$^{3}$ \quad
Eli Waxman$^{2}$\\
$^{1}$Georgia Institute of Technology \qquad $^{2}$MetaDialog \qquad $^{3}$Infuse Inc.
}

\begin{document}
\maketitle
\begin{abstract}
Clinical practice guidelines are long, multimodal documents whose branching recommendations are difficult to convert into executable clinical decision support (CDS), and one-shot parsing often breaks cross-page continuity.
Recent LLM/VLM extractors are mostly local or text-centric, under-specifying section interfaces and failing to consolidate cross-page control flow across full documents into one coherent decision graph.
We present a decomposition-first pipeline that converts full-guideline evidence into an executable clinical decision graph through topology-aware chunking, interface-constrained chunk graph generation, and provenance-preserving global aggregation.
Rather than relying on single-pass generation, the pipeline uses explicit entry/terminal interfaces and semantic deduplication to preserve cross-page continuity while keeping the induced control flow auditable and structurally consistent.
We evaluate on an adjudicated prostate-guideline benchmark with matched inputs and the same underlying VLM backbone across compared methods.
On the complete merged graph, our approach improves edge and triplet precision/recall from $19.6\%/16.1\%$ in existing models to $69.0\%/87.5\%$, while node recall rises from $78.1\%$ to $93.8\%$.
These results support decomposition-first, auditable guideline-to-CDS conversion on this benchmark, while current evidence remains limited to one adjudicated prostate guideline and motivates broader multi-guideline validation.
\end{abstract}
    
\ifthenelse{\equal{\compileArxiv}{true}}{
  \fancypagestyle{firststyle}
  {
     \fancyhead{}
     \lhead{Accepted as a conference paper at CVPR 2026 - MedReasoner Workshop.}
     \renewcommand{\headrulewidth}{0pt}
  }
  \thispagestyle{firststyle}
}{}
\section{Introduction}
\label{sec:intro}
Clinical practice guidelines (CPGs) are foundational to evidence-based medicine, but translating these dense, multi-page documents into actionable clinical decision support (CDS) systems remains a significant bottleneck. Historically, the clinical informatics community has studied how to transform narrative CPGs into computable decision logic. Early \emph{computer-interpretable guideline} (CIG) formalisms, including GLIF \cite{ohnomachado1998glif,boxwala2004glif3}, represented guideline recommendations as shareable step-wise logic. Complementary document-centric frameworks like GEM \cite{gershkovich2001implementation,shiffman2000gem} preserved structural provenance, while rule- and workflow-oriented representations such as Arden Syntax \cite{hripcsak1994rationale,samwald2012arden}, PROforma \cite{sutton2003proforma}, and Asbru \cite{miksch1997asbru} modeled richer workflow and temporal intents. Subsequent systems like SAGE \cite{tu2007sage} emphasized integration with patient data. Despite establishing a foundation for executable clinical workflows, these early frameworks lacked the ability to automatically construct executable decision graphs directly from raw, long, and heterogeneous guideline documents.

\begin{figure}[]
    \centering
    \includegraphics[scale=1]{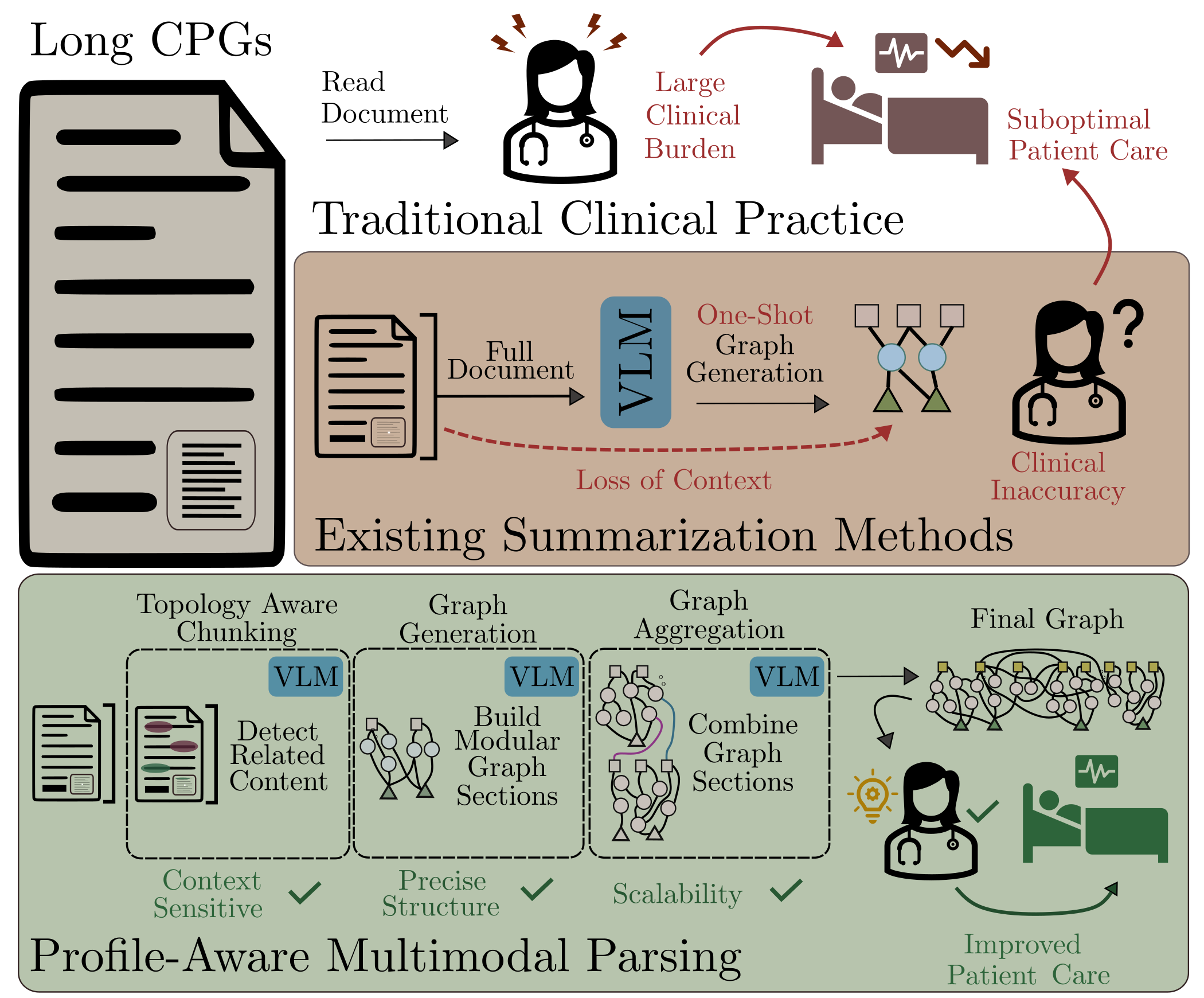}
    \caption{Overview of our profile-aware multimodal parsing framework. Unlike traditional practice and one-shot VLM summarization, our method uses topology-aware chunking, modular graph generation, and graph aggregation to preserve context and structure, yielding a scalable final graph for improved patient care.}
    \label{fig:DGPM}
\end{figure}

\begin{figure*}[]
    \centering
    \includegraphics[scale=1]{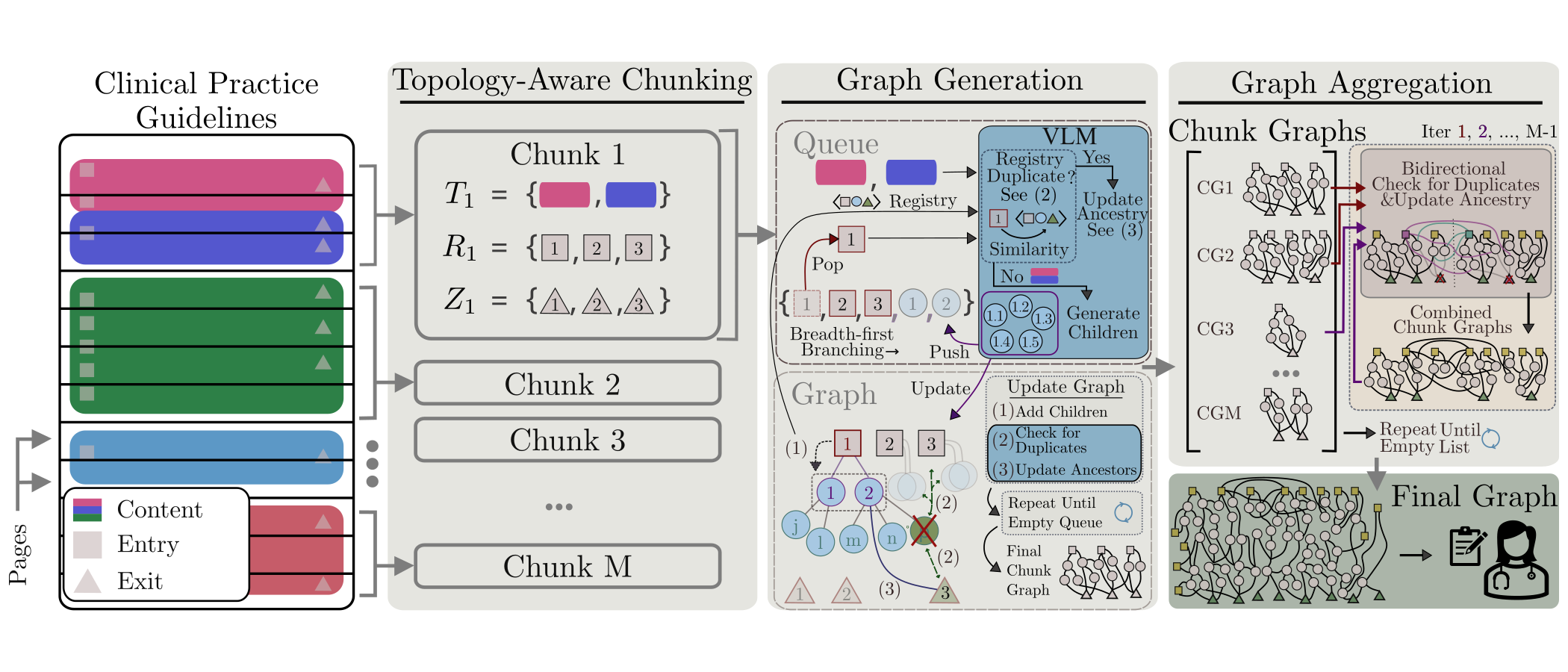}
    \caption{Our detailed pipeline. Long CPGs are split into topology-aware chunks, each chunk graph is built via queue-based VLM expansion (with duplicate and ancestry updates), and all chunk graphs are iteratively merged into a final graph.}
    \label{fig:DGPM}
\end{figure*}

Recent advancements in large language models (LLMs) and vision-language models (VLMs) have attempted to automate this by reframing guideline understanding as a structured extraction task. However, they rely on global parsing strategies that are optimized for short-document contexts or isolated text snippets. They do not provide a principled way to scale graph extraction across long documents where critical branching logic is distributed across complex layouts, and tables, and multi-page text. Hence, their performance are neither scalable nor transferable to long document context graphs. Thus, reliably parsing decision graphs that spans the whole document for long contexts remains as a main challenge.

To address this problem, we introduce a scalable framework that handles long-document graph extraction via topology-aware chunking, graph parsing, and global aggregation. First, our chunking mechanism enables scalability through computation-constrained entry-exit span detection, cross-page multimodal relevant context classification, and canonical node representation. This isolates coherent decision segments without losing the global narrative. Second, our graph parser provides a principled method for iterative node generation, systematically branching out from defined entry nodes toward exit nodes. Finally, our aggregation stage utilizes a semantic matching-based deduplication and stitching method—acting as a retriever—to consolidate the isolated chunks into a single, globally consistent decision graph. An overview of the proposed framework is presented in Figure \ref{fig:DGPM}.

A core distinction of our work is that we do not blindly prompt VLMs for one-shot graph generation. Instead, we rigorously decompose the complex long-document graph parsing problem into targeted sub-problems. Within our pipeline, VLMs are deliberately orchestrated to act as named entity recognizers (NER), boundary detectors, classifiers for structured output logic, and semantic rerankers. By assigning VLMs these specific roles to guide a step-by-step graph extraction, our approach offers a new and fully auditable way to process complex clinical texts.

\textbf{Our contributions} are: \emph{(i)} we propose a topology-aware, profile-guided chunking strategy for full guideline documents that isolates decision-relevant regions while preserving cross-page continuity via canonical entry/exit interfaces; \emph{(ii)} we parse each chunk into a decision subgraph by iteratively branching from the specified entry toward terminal nodes under interface constraints, improving structural consistency across chunks; and \emph{(iii)} we perform embedding-based node deduplication and entity merging with provenance-preserving edge rewiring to stitch chunks into a single consolidated executable decision graph for downstream conformance-oriented CDS evaluation.
\section{Related Work}
\label{sec:related_work}

\paragraph{LLM-based parsing.} Recent work reframes guideline understanding as structured extraction with large models, usually producing decision trees from text. Broader evidence that LLMs encode substantial clinical knowledge motivates their use for guideline interpretation and decision support \cite{singhal2023large}. Text2MDT \cite{zhu2024text2mdt} introduced benchmarked medical decision-tree extraction with multi-level metrics (triplets, nodes, and paths). Generative rule-extraction methods also use constrained sequence formats—linearized representations such as JSON templates—to improve syntactic correctness and make outputs easier to parse \cite{he-etal-2024-generative}. Follow-up systems reduced free-form noise via two-stage generation (if--else scaffold first, then node filling) and reported stronger structural fidelity on curated datasets \cite{hou2025decision}. Related systems also use extracted trees as executable substrates for downstream CDS and vignette-based adherence evaluation, including MedDM \cite{li2023meddm}, binary-tree prompting frameworks, and agentic pipelines such as CPGPrompt \cite{shawi2025leveraging, deng2026cpgprompt}. 

Despite progress, this line of work remains predominantly text-centric: it is typically evaluated on relatively small documents or isolated sections, and it often assumes that the relevant decision logic is explicitly verbalized in prose. In practice, many guidelines distribute key branching logic across tables, figures, and other layout-encoded structures, which can be lost when parsing from plain text alone—making end-to-end extraction brittle on long, multi-page documents with complex formatting. We address this gap by going beyond text-only extraction by grounding graph induction in multimodal page evidence and maintaining cross-page continuity via explicit entry/exit interfaces.

\paragraph{VLM-based multimodal parsing.} In parallel, multimodal document understanding has advanced DocVQA-style reasoning \cite{mathew2021docvqa}, layout-aware parsing with document pretraining \cite{xu2020layoutlm,xu2021layoutlmv2,huang2022layoutlmv3}, and OCR-free document modeling \cite{kim2022ocr}, alongside benchmarks for complex layout and flowchart understanding \cite{zhong2019publaynet,pan2407flowlearn}. These lines of work show that combining textual content with visual-layout cues improves robustness on complex page designs (e.g., multi-column sections, nested tables, and flowchart-like regions) \cite{xu2021layoutlmv2,huang2022layoutlmv3,kim2022ocr}.

However, most multimodal document-understanding systems are framed as question answering or field extraction, rather than end-to-end induction of executable clinical decision graphs from full guidelines. As a result, they often do not produce globally consistent, auditable control-flow structures suitable for downstream CDS. Recent evaluations also report stability issues even with improved perceptual capability, motivating explicit, inspectable multi-stage pipelines over single-pass generation \cite{hsu2025extracting}. Rather than QA/field extraction, we target full-document executable decision-graph induction with interface-constrained iterative expansion from entry to exit nodes.

\paragraph{Long-document parsing.} LLM-assisted decision-graph construction has also been explored for automated and heterogeneous-document settings, including AutoKG \cite{chen2023autokg} and Docs2KG \cite{sun2024docs2kg}. Long-document graph-construction pipelines commonly decompose inputs into local segments, build intermediate structures, and merge them globally to reduce long-context failure modes \cite{edge2024local}. A recurring design pattern is candidate retrieval followed by semantic verification and canonicalization, as exemplified by Extract--Define--Canonicalize \cite{zhang2024extract}. This retrieve--verify--merge pattern is closely related to classical and neural entity-resolution ideas, including Fellegi--Sunter \cite{fellegi1969theory} style linkage and learned pairwise matching (e.g., Ditto \cite{li2020deep}), for reconciling near-duplicate nodes. Search-based reasoning frameworks also motivate explicit state expansion and revision rather than latent one-shot generation \cite{yao2023tree_of_thoughts,besta2024graph_of_thoughts}. 

In guideline informatics, prior CIG research focused on making recommendations computable and deployable in CDS systems \cite{peleg2013computer}. Recent LLM-guideline methods focus on extracting executable decision structures and using them in downstream adherence workflows \cite{shawi2025leveraging,deng2026cpgprompt}.

A key limitation across these strands is that long-document graph methods are mostly designed for general document intelligence, while guideline extraction methods typically emphasize local structure induction over full-document graph consolidation. As a result, prior systems often under-specify interface constraints (entry/terminal consistency), cross-chunk deduplication, and provenance-aware merge operations needed to produce a single auditable executable graph from long multimodal guidelines. In contrast, We contribute guideline-specific global consolidation: interface-consistent chunk stitching with embedding-based deduplication and provenance-preserving edge rewiring into a single auditable graph.

\section{Method}
\label{sec:method}

\begin{algorithm}[t]
\caption{\textsc{Chunk Generation}}
\label{alg:chunking}
\begin{algorithmic}[1]
\Require Document pages $\{P_i\}_{i=1}^n$, header length $h$, soft chunk budget $L$
\Ensure Chunk set $\mathcal{P}=\{(T_j,R_j,Z_j)\}_{j=1}^m$

\State $(M,ct) \gets \textsc{ExtractGuidelineProfile}(\{P_i\}_{i=1}^{h})$
  
\Comment{$M$: metadata; $ct$: scope context}

\ForAll{$i \in \{1,\dots,n\}$ \textbf{in parallel}}
  \State $\tau_i \gets \textsc{ClassifyPage}(P_i, M)$
  
  \Comment{$\tau_i \in \{\textsc{Core},\textsc{Auxiliary}\}$}
\EndFor

\State $I \gets \textsc{Sorted}(\{ i : \tau_i=\textsc{Core}\})$
\State $\mathcal{H} \gets \textsc{ContiguousRuns}(I)$
\Comment{split core pages into maximal consecutive runs}

\State $\mathcal{P} \gets [\ ]$

\ForAll{run $H \in \mathcal{H}$}
  \State $ctx \gets ct$ \Comment{running narrative memory}
  \State $B \gets [\ ]$ \Comment{buffer of pages in current chunk}

  \For{$t \gets 1$ \textbf{to} $|H|$}
    \State $i \gets H[t]$
    \State $P^{+} \gets \begin{cases}
        P_{H[t+1]} & t < |H|\\
        \bot & t = |H|
    \end{cases}$

    \State $\textit{cut} \gets \textsc{PredictBoundary}(B, P_i, P^{+}, ctx, L)$
    \State $B.\textsc{append}(P_i)$

    \If{$\textit{cut}$ \textbf{or} $t=|H|$}
	      \State $(d,R,Z,K,ctx') \gets \textsc{Build}(B, P^{+}, ctx)$
	      \Comment{$d$: description;\ $R/Z$: entry/terminal labels;\ $K$: carry-forward pages}

	      \State $(R,Z) \gets \textsc{RefineNodes}(B, d, R, Z)$

	      \State $T \gets \textsc{AssembleContext}(M, d, B, ctx')$
	      \Comment{includes supporting material in $T$}

	      \State $\mathcal{P}.\textsc{append}((T,R,Z))$

      \State $B \gets \textsc{CarryForwardPages}(B, K)$
      \State $ctx \gets ctx'$
    \EndIf
  \EndFor
\EndFor

\State \Return $\mathcal{P}$
\end{algorithmic}
\end{algorithm}

Given a guideline document represented as an ordered sequence of pages $\{P_i\}_{i=1}^n$, we convert the document into an explicit, interpretable \emph{decision graph} $G=(V,E)$ capturing step-by-step clinical reasoning, where $V$ is the set of nodes and $E$ is the set of directed labeled edges. Each node $v\in V$ corresponds to a clinical state/decision/recommendation. Each edge tuple $(u,\ell,v)\in E$ denotes a transition from source node $u$ to destination node $v$ under condition/label $\ell$.

Directly extracting a global graph from an entire guideline is infeasible due to long context length and the presence of non-decision material (references, appendices, acknowledgements). Therefore, we use a three-stage pipeline:
\begin{enumerate}
  \item \textbf{Chunk generation} to isolate coherent ``core'' decision segments
within a soft budget (Alg.~\ref{alg:chunking});
  \item \textbf{Chunk-level graph generation} using a queue-based expansion with intra-chunk deduplication (Alg.~\ref{alg:build-subgraph});
  \item \textbf{Global aggregation} that merges all chunk graphs into a single graph via cross-chunk deduplication and edge rewiring (Alg.~\ref{alg:aggregate}).
\end{enumerate}

Throughout, key semantic steps are implemented using a prompted vision-language model (VLM). When page images are available, we provide the rendered page image together with extracted text; otherwise we run the same prompt in text-only mode. 

\begin{table*}[h]
\centering
\caption{Comparison of directed-graph (DG) construction approaches. 
Our method constructs a persistent, canonicalized decision DG from long documents 
with BFS-based expansion and semantic deduplication.}
\label{tab:method_comparison}
\small
\begin{tabular}{lccccc}
\toprule
\textbf{Method} 
& \textbf{Long} 
& \textbf{Persistent} 
& \textbf{DG Type} 
& \textbf{BFS} 
& \textbf{Intra-Node} \\
& \textbf{Document} 
& \textbf{Structure} 
&  
& \textbf{Expansion} 
& \textbf{Dedup.} \\
\midrule
Tree-of-Thought (ToT) 
& \xmark & \xmark & Ephemeral search & \cmark & \xmark \\

Graph-of-Thought (GoT) 
& \xmark & \xmark & Inference-time reasoning & \xmark & \xmark \\

Med-PaLM
& \xmark & \xmark & None (text output) & \xmark & \xmark \\

Doc2KG
& \cmark & \cmark & Relational triple DG & \xmark & Limited \\

AutoKG 
& \cmark & \cmark & Relational triple DG & \xmark & Limited \\

\midrule
\textbf{Ours} 
& \cmark & \cmark & Canonicalized decision DG & \cmark & \cmark \\
\bottomrule
\end{tabular}
\end{table*}

\subsection{Document Representation}

For each page index $i$, we construct a unified page object $P_i$ from: (1) an optional rendered page image $I_i$ (if available), and (2) page text $x_i$, obtained from the portable document format (PDF) text layer when present, otherwise via optical character recognition (OCR). Concretely, $P_i=(I_i,x_i)$, which becomes $(\varnothing,x_i)$ when no image is available, and the chunk-generation stage operates directly on $\{P_i\}_{i=1}^n$. This design allows the same pipeline to operate on born-digital PDFs (text-only), scanned PDFs (via OCR), and mixed documents (image + OCR). Accordingly, components that rely on layout cues (e.g., flowchart/table detection or section-boundary identification) consume both $(I_i, x_i)$, whereas text-only components consume $x_i$.

\subsection{Chunk Generation (Alg. \ref{alg:chunking})}

We first derive a document-level profile $M$ from the first $h$ pages. In addition to structured metadata (e.g., title and guideline code), this step produces a compact scope context $ct$ that captures the intended population and clinical focus. We retain $ct$ throughout chunking so local segmentation decisions remain consistent with the global guideline intent. We denote by $L$ the soft chunk-length budget used by boundary prediction.

Each page is then labeled as $\tau_i\in\{\textsc{Core},\textsc{Auxiliary}\}$ using a prompted VLM. Core pages contain actionable decision content (algorithms, criteria, recommendations, flowcharts), whereas auxiliary pages contain non-decision material such as references, author lists, and administrative text. Because this decision is page-local, labeling is parallelized across pages.

Let $I=\{i:\tau_i=\textsc{Core}\}$ be the set of core-page indices. We partition $I$ into maximal consecutive runs $\mathcal{H}$ to prevent chunk construction from spanning long auxiliary gaps, which improves local semantic continuity.

For each run $H\in\mathcal{H}$, chunks are built incrementally with a page buffer $B$ and a running memory $ctx$, initialized as $ct$. At page $P_i$, we compute a boundary decision from the tuple $(B, P_i, P^+, ctx, L)$, where $P^+$ is a one-step lookahead page and $L$ is a soft token/length budget. Including $P^+$ reduces boundaries that would separate section headers from supporting text or split multi-page tables/figures.

When a boundary is triggered (or the run ends), the buffered pages are finalized into one chunk. For chunk index $j$, we obtain a short chunk description $d_j$, \emph{entry} labels $R_j$, \emph{terminal} labels $Z_j$, carry-forward pages $K_j$ for inter-chunk continuity, and an updated memory $ctx'_j$. The interface labels $(R_j,Z_j)$ are then normalized and checked for textual support, and the final chunk context $T_j$ (profile-grounded local evidence and running context) is assembled from $(M,d_j,B,ctx'_j)$ for downstream graph generation.

The output of this stage is a sequence of chunk triplets $\mathcal{P}=\{(T_j,R_j,Z_j)\}_{j=1}^m$, where $m$ is the number of chunks, $R_j$ is the set of entry/root node labels used to initialize graph expansion, and $Z_j$ is the fixed set of terminal node labels for that chunk.

\subsection{Chunk-Level Graph Generation (Alg.~\ref{alg:build-subgraph})}
Given chunk index $j$ and its tuple $(T_j,R_j,Z_j)$, we construct a local graph $G_j=(V_j,E_j)$ via breadth-first expansion from entry nodes while enforcing intra-chunk consistency. The terminal set $Z_j$ is provided as an input interface and is treated as fixed for the chunk: terminal nodes are initialized in $V_j$ up front, and no additional terminal nodes are introduced during expansion. The queue is initialized only with root nodes in $R_j$, which are provided by the chunking mechanism. Each queued item stores a node candidate $u$ with incoming context $\alpha\in\{\bot,(a,e)\}$, where $\bot$ denotes no incoming parent-edge context, $a$ is the immediate ancestor of $u$, and $e$ is the label on edge $a\xrightarrow{e}u$.
For nearest-neighbor retrieval operations, we use a fixed candidate count $k$ (i.e., top-$k$ cosine neighbors).

For each dequeued candidate, we retrieve top-$k$ in-chunk neighbors $S$ by embedding similarity using the pair $(u,V_j)$, where $V_j$ already contains all terminal nodes in $Z_j$. Semantic equivalence is then decided from the triplet $(u,\alpha,S)$ with a prompted VLM verifier. If a duplicate $u^\star$ is found, incoming edges are redirected to $u^\star$; otherwise, $u$ is registered as a new \emph{non-terminal} node. In particular, if a generated candidate corresponds to a terminal state, it is merged into an existing $z\in Z_j$ rather than added as a new terminal.
\begin{algorithm}[t]
\caption{\textsc{Graph Generation}}
\label{alg:build-subgraph}
\begin{algorithmic}[1]
\Require Chunk text/context $T_j$, root nodes $R_j$, terminal nodes $Z_j$
\Ensure Chunk graph $G_j = (V_j,E_j)$

\State $V_j \gets Z_j$
\State $E_j \gets \varnothing$
\State $\mathcal{Q} \gets$ empty queue

\ForAll{$r \in R_j$}
  \State \textsc{Enqueue}$(\mathcal{Q}, (r,\bot))$
\EndFor

\While{\textsc{NotEmpty}$(\mathcal{Q})$}
  \State $(u,\alpha) \gets$ \textsc{Dequeue}$(\mathcal{Q})$
  \Comment{$\alpha=\bot$ or $\alpha=(a,e)$ meaning $a \xrightarrow{e} u$}
  \State $S \gets \textsc{CosineCandidates}(u, V_j)$
    \Comment{top-$k$ by embedding similarity}
  \State $u^\star \gets \textsc{FindDuplicate}(u,\alpha,S)$
    \If{$u^\star \neq \varnothing$}
    \If{$\alpha \neq \bot$}
        \State $(a,e) \gets \alpha$
        \State \textsc{RedirectAncestorEdge}$(E_j,\ (a,e,u)\rightarrow(a,e,u^\star))$
    \EndIf
    \State \textbf{continue}
    \Else
		    \State \textsc{RegisterNode}$(V_j, E_j, u, \alpha)$
	  \EndIf

		  \State $\mathcal{C} \gets \textsc{GenerateChildren}(u,\alpha,T_j)$
  \Comment{$\mathcal{C}=\{(v,e_{uv}) : u \xrightarrow{e_{uv}} v\}$}
  \ForAll{$(v,e_{uv})\in \mathcal{C}$}
    \State \textsc{Enqueue}$(\mathcal{Q}, (v,(u,e_{uv})))$
  \EndFor
\EndWhile

\State \Return $G_j=(V_j,E_j)$
\end{algorithmic}
\end{algorithm}

Each registered non-terminal node is then expanded by generating clinically valid successors from node context and chunk context, i.e., from $(u,\alpha,T_j)$. The model outputs pairs $\{(v,e_{uv})\}$, where $e_{uv}$ is the transition condition from $u$ to $v$. Each successor is enqueued with context $(u,e_{uv})$, and expansion continues until the queue is exhausted, yielding a chunk graph whose terminal nodes are exactly the input set $Z_j$.

\begin{algorithm}[h!]
\caption{\textsc{Global Aggregation Across Chunks}}
\label{alg:aggregate}
\begin{algorithmic}[1]
\Require Document chunks $\mathcal{P}=\{(T_j,R_j,Z_j)\}_{j=1}^m$
\Ensure Global merged decision graph $G=(V,E)$

\State $V \gets \varnothing$, $E \gets \varnothing$
\For{$j \gets 1$ \textbf{to} $m$}
    \State $G_j=(V_j,E_j) \gets \textsc{GenerateGraph}(T_j,R_j,Z_j)$ \hfill \Comment{Alg.~\ref{alg:build-subgraph}}
    \State $V \gets V \cup V_j$
    \State $E \gets E \cup E_j$
    \ForAll{$v \in V_j$}
        \State $\mathrm{orig}(v) \gets j$
    \EndFor
\EndFor
\State $G \gets (V,E)$

\vspace{0.5em}
\State $\mathcal{Q} \gets$ empty queue \Comment{seed with interface nodes only: $R_j$ and $Z_j$}
\For{$j \gets 1$ \textbf{to} $m$}
    \ForAll{$r \in R_j$}
        \State \textsc{Enqueue}$(\mathcal{Q}, r)$
    \EndFor
    \ForAll{$z \in Z_j$}
        \State \textsc{Enqueue}$(\mathcal{Q}, z)$
    \EndFor
\EndFor

\vspace{0.5em}
\While{\textsc{NotEmpty}$(\mathcal{Q})$}
    \State $x \gets$ \textsc{Dequeue}$(\mathcal{Q})$

    \State $\mathcal{A} \gets \textsc{GetAncestors}(G,x)$ 
      \Comment{$\mathcal{A}=\{(a,e): a \xrightarrow{e} x\}$}

    \State $C \gets \{y \in V : \mathrm{orig}(y) \neq \mathrm{orig}(x)\}$
      \Comment{exclude same-origin nodes}

    \State $S \gets \textsc{CosineCandidates}(x, C)$
      \Comment{top-$k$ by embedding similarity}
    \State $x^\star \gets \textsc{FindDuplicate}(x,\mathcal{A},S)$

    \If{$x^\star \neq \varnothing$}
        \State $(p,s) \gets \textsc{ChoosePrimarySecondary}(x,x^\star)$
        \Comment{prefer non-terminal; then earlier chunk index}

        \ForAll{$(u,\ell,s) \in E$}
            \State \textsc{RedirectAncestorEdge}$(E,\ (u,\ell,s)\rightarrow(u,\ell,p))$
        \EndFor

        \ForAll{$(s,\ell,v) \in E$}
            \State \textsc{RedirectSuccessorEdge}$(E,\ (s,\ell,v)\rightarrow(p,\ell,v))$
        \EndFor

        \If{$s \in \mathcal{Q}$}
            \State \textsc{Remove}$(\mathcal{Q}, s)$
        \EndIf
    \EndIf
\EndWhile

\State \Return $G=(V,E)$
\end{algorithmic}
\end{algorithm}

\subsection{Global Aggregation Across Chunks (Alg.~\ref{alg:aggregate})}

Once all chunks are generated, we merge their corresponding chunk graphs into a single document-level graph while resolving duplicate nodes at chunk interfaces. For each chunk $j$, we first construct its local graph $G_j=(V_j,E_j)$ from $(T_j,R_j,Z_j)$, then form the global union $V=\cup_j V_j$ and $E=\cup_j E_j$. We also retain provenance $\mathrm{orig}(v)=j$ for each node, indicating the chunk from which $v$ originates.

Cross-chunk duplicates are most frequent around interface nodes, so we initialize a queue with all roots $R_j$ and terminals $Z_j$ from every chunk. For each dequeued node $x$, we collect its ancestor-edge context $\mathcal{A}=\{(a,e):a\xrightarrow{e}x\}$ and restrict duplicate search to cross-chunk candidates $C=\{y\in V:\mathrm{orig}(y)\neq\mathrm{orig}(x)\}$. Using $(x,C)$, we retrieve top-$k$ semantic neighbors $S$, and then evaluate equivalence from the triplet $(x,\mathcal{A},S)$ with a prompted VLM verifier.

If a duplicate is detected, we select a primary node $p$ and a secondary node $s$, preferring non-terminal nodes and breaking ties by chunk order. We then merge by rewiring all incoming and outgoing edges incident to $s$ toward $p$, and remove stale queue entries for $s$ when needed. This yields a globally consolidated graph with reduced interface-level redundancy.

\section{Experiments}
\label{sec:experiments}
This section validates whether the proposed pipeline improves both local extraction fidelity and global graph consolidation under controlled, matched conditions. We first define the evaluation units and adjudicated references, then compare methods with the same inputs, normalization protocol, and underlying VLM backbone to isolate graph-construction effects from backbone effects. We report quantitative node/edge/triplet precision--recall with traceable supported-over-total counts and complement those metrics with qualitative structural analysis to inspect path-level behavior and failure modes.
\subsection{Experimental Setup}
We evaluate long-document clinical decision-graph extraction on a single prostate clinical practice guideline \cite{nccn_prostate_cancer_4_2024}. Following the chunking design in Sec.~\ref{sec:method}, we define six evaluation units: five chunk-level graphs $G_1,\dots,G_5$ and one merged complete graph $G_{\text{all}}$. For each method, outputs are normalized into a common directed labeled graph interface $G=(V,E)$, where $V$ denotes decision/clinical-state nodes and $E$ denotes directed condition-labeled transitions. To ensure parity, all methods are rerun on the same document inputs and normalized with the same post-processing interface before scoring.

Ground-truth references for each unit are manually curated and adjudicated by human reviewers. These adjudicated references are used for node-, edge-, and triplet-level evaluation. We report both percentages and supported-over-total counts (S/T), so each score can be traced back to matched items and denominator size.

\subsection{Compared Methods and Metric Protocol}
Table~\ref{tab:method_comparison} positions our method against representative alternatives, highlighting long-document handling, persistence of graph structure, and deduplication behavior. In quantitative comparisons, we evaluate against Doc2KG \cite{sun2024docs2kg} and AutoKG \cite{chen2023autokg} under matched inputs and normalization. All compared methods use the same underlying VLM backbone, so observed differences are attributable to graph-construction strategy rather than model choice.

\begin{table*}[]
    \centering
    \caption{Precision and recall (\%) for decision graph extraction evaluated on nodes, edges, and full triplets (node--edge--node). ``S/T'' denotes supported-over-total counts used to compute each percentage. Best results per row group are \textbf{bold}.}
    \label{tab:kg_accuracy}
    \vspace{2pt}
    \small
    \renewcommand{\arraystretch}{1.05}
    \setlength{\tabcolsep}{4.5pt}
    \begin{tabular}{@{}ll rr rr rr rr rr rr@{}}
        \toprule
        & & \multicolumn{4}{c}{\textbf{Nodes}} & \multicolumn{4}{c}{\textbf{Edges}} & \multicolumn{4}{c}{\textbf{Triplets}} \\
        \cmidrule(lr){3-6} \cmidrule(lr){7-10} \cmidrule(lr){11-14}
        & & \multicolumn{2}{c}{Prec.} & \multicolumn{2}{c}{Rec.} & \multicolumn{2}{c}{Prec.} & \multicolumn{2}{c}{Rec.} & \multicolumn{2}{c}{Prec.} & \multicolumn{2}{c}{Rec.} \\
        \cmidrule(lr){3-4} \cmidrule(lr){5-6} \cmidrule(lr){7-8} \cmidrule(lr){9-10} \cmidrule(lr){11-12} \cmidrule(lr){13-14}
        Graph & Method & \multicolumn{1}{c}{\%} & \multicolumn{1}{c}{\footnotesize S/T} & \multicolumn{1}{c}{\%} & \multicolumn{1}{c}{\footnotesize S/T} & \multicolumn{1}{c}{\%} & \multicolumn{1}{c}{\footnotesize S/T} & \multicolumn{1}{c}{\%} & \multicolumn{1}{c}{\footnotesize S/T} & \multicolumn{1}{c}{\%} & \multicolumn{1}{c}{\footnotesize S/T} & \multicolumn{1}{c}{\%} & \multicolumn{1}{c}{\footnotesize S/T} \\
        \midrule
        \multirow{3}{*}{1} & Doc2KG & 23.5 & 4/17 & 80.0 & 4/5 & 10.3 & 3/29 & \textbf{75.0} & 3/4 & 10.3 & 3/29 & \textbf{75.0} & 3/4 \\
         & AutoKG & 50.0 & 5/10 & \textbf{100.0} & 5/5 & 11.1 & 1/9 & 25.0 & 1/4 & 11.1 & 1/9 & 25.0 & 1/4 \\
         & Ours & \textbf{80.0} & 4/5 & 80.0 & 4/5 & \textbf{100.0} & 3/3 & \textbf{75.0} & 3/4 & \textbf{100.0} & 3/3 & \textbf{75.0} & 3/4 \\
        \addlinespace[3pt]
        \multirow{3}{*}{2} & Doc2KG & 27.3 & 3/11 & 30.0 & 3/10 & 0.0 & 0/17 & 0.0 & 0/13 & 0.0 & 0/17 & 0.0 & 0/13 \\
         & AutoKG & 41.7 & 10/24 & \textbf{100.0} & 10/10 & 9.5 & 2/21 & 15.4 & 2/13 & 9.5 & 2/21 & 15.4 & 2/13 \\
         & Ours & \textbf{83.3} & 10/12 & \textbf{100.0} & 10/10 & \textbf{73.3} & 11/15 & \textbf{84.6} & 11/13 & \textbf{66.7} & 10/15 & \textbf{76.9} & 10/13 \\
        \addlinespace[3pt]
        \multirow{3}{*}{3} & Doc2KG & 12.5 & 1/8 & 10.0 & 1/10 & 0.0 & 0/12 & 0.0 & 0/14 & 0.0 & 0/12 & 0.0 & 0/14 \\
         & AutoKG & 45.5 & 10/22 & \textbf{100.0} & 10/10 & 28.6 & 6/21 & 42.9 & 6/14 & 19.0 & 4/21 & 28.6 & 4/14 \\
         & Ours & \textbf{75.0} & 9/12 & 90.0 & 9/10 & \textbf{100.0} & 14/14 & \textbf{100.0} & 14/14 & \textbf{92.9} & 13/14 & \textbf{92.9} & 13/14 \\
        \addlinespace[3pt]
        \multirow{3}{*}{4} & Doc2KG & 11.1 & 1/9 & 12.5 & 1/8 & 0.0 & 0/13 & 0.0 & 0/12 & 0.0 & 0/13 & 0.0 & 0/12 \\
         & AutoKG & 36.8 & 7/19 & 87.5 & 7/8 & 0.0 & 0/18 & 0.0 & 0/12 & 0.0 & 0/18 & 0.0 & 0/12 \\
         & Ours & \textbf{53.3} & 8/15 & \textbf{100.0} & 8/8 & \textbf{66.7} & 10/15 & \textbf{83.3} & 10/12 & \textbf{40.0} & 6/15 & \textbf{50.0} & 6/12 \\
        \addlinespace[3pt]
        \multirow{3}{*}{5} & Doc2KG & 16.7 & 1/6 & 9.1 & 1/11 & 0.0 & 0/9 & 0.0 & 0/13 & 0.0 & 0/9 & 0.0 & 0/13 \\
         & AutoKG & 47.6 & 10/21 & 90.9 & 10/11 & 19.0 & 4/21 & 30.8 & 4/13 & 14.3 & 3/21 & 23.1 & 3/13 \\
         & Ours & \textbf{55.0} & 11/20 & \textbf{100.0} & 11/11 & \textbf{50.0} & 12/24 & \textbf{92.3} & 12/13 & \textbf{45.8} & 11/24 & \textbf{84.6} & 11/13 \\
        \midrule
        \multirow{3}{*}{\shortstack[l]{Complete Graph}} & Doc2KG & 27.5 & 14/51 & 43.8 & 14/32 & 1.1 & 1/88 & 1.8 & 1/56 & 1.1 & 1/88 & 1.8 & 1/56 \\
         & AutoKG & 56.8 & 25/44 & 78.1 & 25/32 & 19.6 & 9/46 & 16.1 & 9/56 & 19.6 & 9/46 & 16.1 & 9/56 \\
         & Ours & \textbf{57.7} & 30/52 & \textbf{93.8} & 30/32 & \textbf{69.0} & 49/71 & \textbf{87.5} & 49/56 & \textbf{69.0} & 49/71 & \textbf{87.5} & 49/56 \\
        \bottomrule
    \end{tabular}
\end{table*}

For each graph unit, we compute precision and recall on nodes, edges, and triplets (node--edge--node), where triplets capture topological consistency. Precision uses prediction-to-ground-truth matching (supported predictions / total predictions), and recall uses ground-truth-to-prediction matching (supported ground-truth items / total ground-truth items). We report these as both percentages and S/T counts in Table~\ref{tab:kg_accuracy}.

\subsection{Quantitative Results}
On the merged complete graph $G_{\text{all}}$, our method shows clear structural gains over AutoKG in Table~\ref{tab:kg_accuracy}: node recall improves by $+15.7$ points ($93.8\%$ vs. $78.1\%$), edge precision by $+49.4$ points ($69.0\%$ vs. $19.6\%$), and edge recall by $+71.4$ points ($87.5$ vs. $16.1$). Triplet precision/recall show the same $+49.4$/$+71.4$ improvements ($69.0\%/87.5\%$ vs. $19.6\%/16.1\%$), indicating better preservation of executable transition structure after global aggregation.

Across chunk-level units $G_1$--$G_5$, our method is consistently strong: it outperforms the best baseline in $25/30$ metric cells and ties in $3/30$. This pattern suggests the gains are not confined to a single chunk and remain stable across local decision segments. One nuance is node recall on $G_1$ and $G_3$, where AutoKG reaches $100.0\%$ and ours is lower ($80.0\%$ and $90.0\%$). However, the structural metrics that determine executable control flow are substantially better for ours (e.g., $G_3$ edges: $100.0\%/100.0\%$ precision/recall vs. $28.6\%/42.9\%$ for AutoKG; triplets: $92.9\%/92.9\%$ vs. $19.0\%/28.6\%$), supporting higher graph fidelity despite isolated recall trade-offs.

\subsection{Qualitative Analysis}
\begin{figure*}[t]
    \centering
    \includegraphics[scale=0.88]{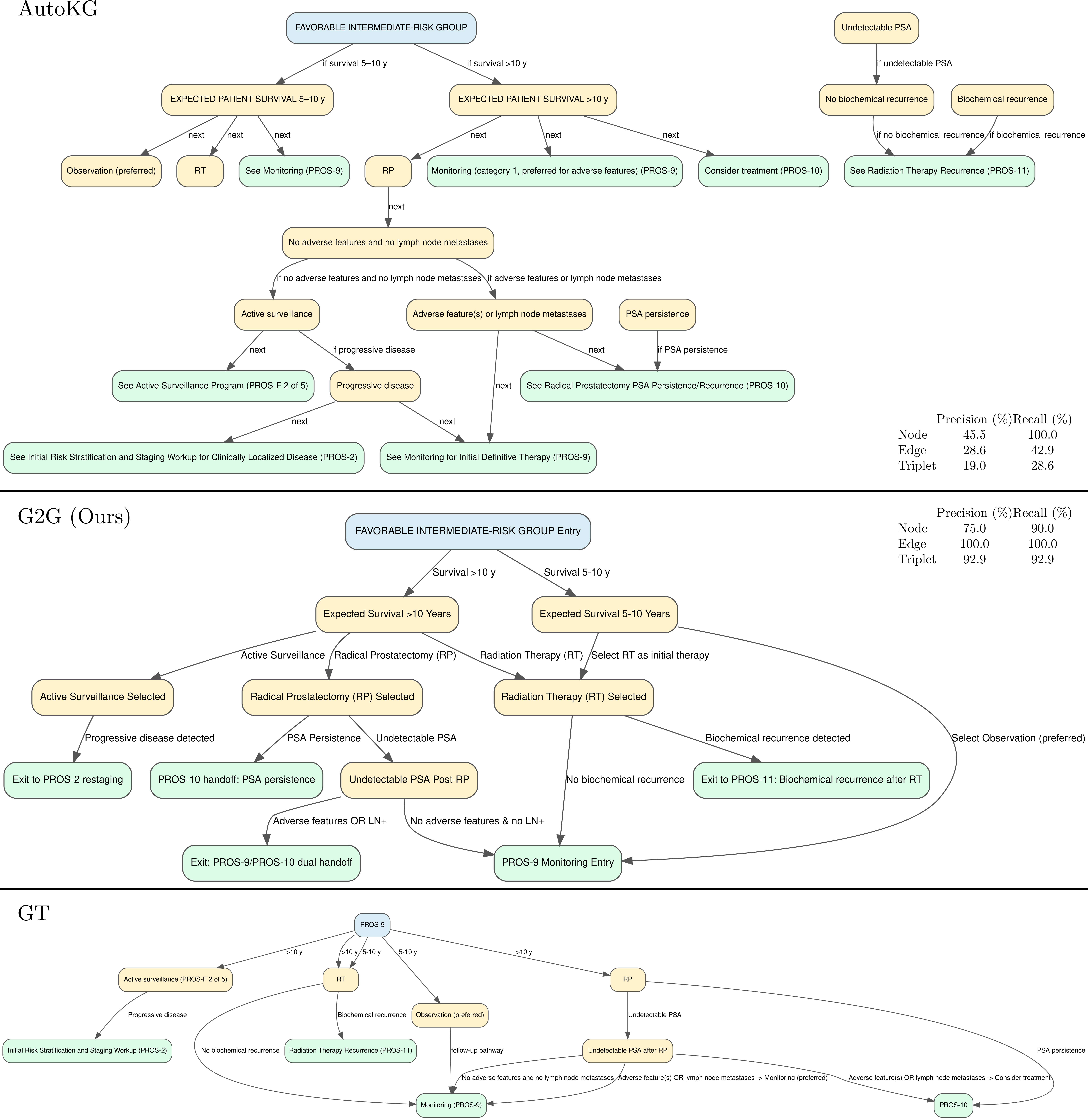}
    \caption{Qualitative comparison on one representative decision module. (A) AutoKG baseline output, (B) our output, and (C) adjudicated ground-truth graph. Our method better preserves path continuity and branching fidelity, with fewer spurious/fragmented transitions.}
    \label{fig:qualitative}
\end{figure*}

Figure~\ref{fig:qualitative} provides a structural comparison on a representative module. Panel A shows the AutoKG baseline output, Panel B shows our output, and Panel C shows the adjudicated ground truth. This side-by-side view complements the tabular metrics by exposing path-level behavior directly.

Relative to the baseline, our graph preserves longer path continuity, cleaner branch assignments, and fewer fragmented or spurious transitions. In particular, Panel B more faithfully reproduces Panel C by preserving the treatment-choice branches (active surveillance/RP/RT) and their downstream follow-up and recurrence links, consistent with the higher edge/triplet precision and recall in Table~\ref{tab:kg_accuracy}.

\subsection{Discussion and Limitations}
These results show that our pipeline produces higher-fidelity long-document decision graphs in this benchmark, especially on edges and triplets. The per-chunk and merged evaluations expose both local parsing quality and global consolidation behavior within the same framework. The main remaining limitation is benchmark breadth: due to adjudicated ground-truth availability, evaluation is currently on one guideline, and expanding annotated guidelines is the next step to confirm cross-guideline generalization.

\section{Conclusion}
\label{sec:conclusion}

We presented a scalable framework for extracting executable clinical decision graphs from long clinical guideline documents. The approach combines topology-aware chunking with explicit entry/terminal interfaces, iterative chunk-level graph generation with semantic deduplication, and provenance-preserving global aggregation into a single consolidated graph. On our adjudicated prostate-guideline benchmark, using the same underlying VLM backbone across methods, our system achieves the strongest structural fidelity on the merged graph, improving edge precision/recall from $19.6\%/16.1\%$ (AutoKG) to $69.0\%/87.5\%$, with matching gains on node--edge--node triplets. Qualitative analysis further shows cleaner treatment/follow-up branching and fewer fragmented transitions relative to baseline outputs. Overall, these results establish the effectiveness of decomposition-first long-document graph induction for auditable guideline-to-CDS conversion. Beyond demonstrating end-to-end performance, they show that VLMs can reliably handle low-level subtasks and point to a promising direction: pairing compute-efficient, task-specialized models with targeted training to further improve reliability and efficiency. 
{
    \small
    \bibliographystyle{ieeenat_fullname}
    \bibliography{main}
}


\end{document}